# A network of spiking neurons for computing sparse representations in an energy efficient way

**Tao Hu**[1]
*hut@janelia.hhmi.org*

**Alexander Genkin**[2]
*alexgenkin@iname.com*

**Dmitri B. Chklovskii**[1]
*mitya@janelia.hhmi.org*

[1]*Howard Hughes Medical Institute, Janelia Farm Research Campus, Ashburn, VA 20147, USA*
[2]*AVG Consulting, Brooklyn, NY, USA*

**Computing sparse redundant representations is an important problem both in applied mathematics and neuroscience. In many applications, this problem must be solved in an energy efficient way. Here, we propose a hybrid distributed algorithm (HDA), which solves this problem on a network of simple nodes communicating via low-bandwidth channels. HDA nodes perform both gradient-descent-like steps on analog internal variables and coordinate-descent-like steps via quantized external variables communicated to each other. Interestingly, such operation is equivalent to a network of integrate-and-fire neurons, suggesting that HDA may serve as a model of neural computation. We show that the numerical performance of HDA is on par with existing algorithms. In the asymptotic regime the representation error of HDA decays with time, *t*, as 1/*t*. HDA is stable against time-varying noise, specifically, the representation error decays as $1/\sqrt{t}$ for Gaussian white noise.**

## 1    Introduction

Many natural signals can be represented as linear combinations of a few feature vectors (or elements) chosen from a redundant (or overcomplete) dictionary. Such representations are called sparse because most dictionary elements enter with zero coefficients. The importance of sparse representations has been long recognized in applied mathematics (Baraniuk, 2007; Chen, Donoho, & Saunders, 1998) and in neuroscience, where electrophysiological recordings (DeWeese, Wehr, & Zador, 2003) and theoretical arguments (Attwell & Laughlin, 2001; Lennie, 2003) demonstrate that most neurons are silent at any given moment (Gallant & Vinje, 2000; Olshausen & Field, 1996, 2004).

In applied mathematics, sparse representations lie at the heart of many important developments. In signal processing, such solutions serve as a foundation for basis pursuit (Chen, et al., 1998) de-noising, compressive sensing (Baraniuk, 2007) and object recognition (Kavukcuoglu et al., 2010). In statistics, regularized multivariate regression algorithms, such as the Lasso (Tibshirani, 1996) or the elastic net (Zou & Hastie, 2005), rely on sparse representations to perform feature subset selection along with coefficient fitting. Given the importance of finding sparse representations, it is not surprising that many algorithms have been proposed for the task (Cai, Osher, & Shen, 2009a, 2009b; Efron, Hastie, Johnstone, & Tibshirani, 2004; Friedman, Hastie, Hofling, & Tibshirani, 2007; Li & Osher, 2009; Xiao, 2010; Yin, Osher, Goldfarb, & Darbon, 2008; Zou & Hastie, 2005). However, most algorithms are designed for CPU architectures and are computationally and energy intensive.

Given the ubiquity of sparse representations in neuroscience, how can neural networks generate sparse representations remains a central question. Building on the seminal work of Olshausen and Field (Olshausen & Field, 1996), Rozell et al. have proposed an algorithm for sparse representations by neural networks called Local Competitive Algorithm (LCA) (Rozell, Johnson, Baraniuk, & Olshausen, 2008). Such algorithm computes a sparse representation on a network of nodes that communicate analog variables with each other. Although a step towards biological realism, the LCA neglects the fact that most neurons communicate using action potentials (or spikes) - quantized all-or-none electrical signals. In principle, analog variables can be represented by firing rates if they are sufficiently high. However, this limit erases the advantage of spiking in terms of energy efficiency, an important consideration in brain design (Attwell & Laughlin, 2001; Laughlin & Sejnowski, 2003). On the other hand, reducing firing rates often reveals their computational inferiority relative to pure analog because of the punctuate nature of spike trains. (Deneve & Boerlin, 2011; Shapero, Brüderle, Hasler, & Rozell, 2011).

In this paper, we introduce an energy efficient algorithm called hybrid distributed algorithm (HDA), which computes sparse redundant representations on the architecture of (Rozell, et al., 2008) using spiking neurons without resorting to the limit of high firing rates. We demonstrate that, despite the punctuate nature of spike trains, such algorithm performs as well as the analog one, thus suggesting that spikes may not detrimentally affect computational capabilities of neural circuits. Moreover, HDA can serve as a plausible model of neural computation because local operations are described by the biologically inspired integrate-and-fire neurons (Dayan & Abbott, 2001; Koch, 1999). Other spiking neuron models have been proposed for sensory integration, working memory (Boerlin & Deneve, 2011) and implementing dynamical systems (Deneve & Boerlin, 2011; Shapero, et al., 2011).

Because spiking communication requires smaller bandwidth, HDA may also be useful for sensor networks, which must discover sparse causes in distributed signals. In particular, large networks of small autonomous nodes are commonly deployed both for civilian and military applications, such as monitoring the motion of tornado or forest fires, tracking traffic conditions, security surveillance in shopping malls and parking facilities, locating and tracking enemy movements, detection of terrorist threats and attacks, (Tubaishat & Madria, 2003). The nodes of such networks use finite-life or slowly charging batteries and, hence, must operate under limited energy budget. Therefore, low-energy computations and limited bandwidth communication are two central design principles of such networks. Because correlations are often present among distributed sensor nodes, computing sparse redundant representations is an important task.

The paper is organized as follows. In §2 we describe the Bregman iteration method for computing sparse representations and briefly introduce two other distributed methods. We then consider a refined Bregman iteration method with coordinate descent modifications ( §3) and continue in §4 by deriving our hybrid distributed algorithm. In §5 we prove the asymptotic performance guarantee of HDA, and demonstrate its numerical performance in §6. Finally, we conclude with the discussion of the advantages of HDA ( §7).

## 2 Problem statement and existing distributed algorithms

A sparse solution $\boldsymbol{u} \in \mathbb{R}^n$ of the equation $\boldsymbol{Au} = \boldsymbol{f}$, where $\boldsymbol{f} \in \mathbb{R}^m$, and wide matrix $\boldsymbol{A} \in \mathbb{R}^{m \times n}$ $(n > m)$ can be found by solving the following constrained optimization problem:

$$\min \ \|\boldsymbol{u}\|_1 \ \text{s.t.} \ \boldsymbol{Au} = \boldsymbol{f}, \qquad (1)$$

which is known as basis pursuit (Chen, et al., 1998). In practical applications, where $\boldsymbol{f}$ contains noise, one typically formulates the problem differently, in terms of an unconstrained optimization problem known as the Lasso (Tibshirani, 1996):

$$\min \tfrac{1}{2}\|Au - f\|_2^2 + \lambda\|u\|_1, \tag{2}$$

where $\lambda$ is the regularization parameter which controls the trade-off between representation error and sparsity. The choice of regularization by $l_1$-norm assures that the problem both remains convex (Bertsekas, 2009; Boyd & Vandenberghe, 2004; Dattorro, 2008) and favors sparse solutions (Chen, et al., 1998; Tibshirani, 1996). In this paper we introduce an energy efficient algorithm that searches for a solution to the constrained optimization problem (1) by taking steps towards solving a small number of unconstrained optimization problems (2). Our algorithm is closest to the family of algorithms called Bregman iterations (Cai, et al., 2009a, 2009b; Osher, Mao, Dong, & Yin, 2010; Yin, et al., 2008), which take their name from the replacement of the $l_1$-norm by its Bregman divergence (Bregman, 1967), $D(u, u^k) = \lambda\|u\|_1 - \lambda\|u^k\|_1 - \langle p^k, u - u^k\rangle$, where $p$ is a sub-gradient of $\lambda\|u\|_1$ (Boyd & Vandenberghe, 2004).. The iterations start with $u^0 = p^0 = 0$ and consist of two steps:

$$u^{k+1} = \operatorname{argmin}_u E$$
$$= \operatorname{argmin}_u \{\tfrac{1}{2}\|Au - f\|_2^2 + \lambda\|u\|_1 - \lambda\|u^k\|_1 - \langle p^k, u - u^k\rangle\}. \tag{3}$$
$$p^{k+1} = p^k - A^T(Au^{k+1} - f). \tag{4}$$

Throughout the paper, we assume that $A$ is column normalized, i.e. if $A_i$ is the $i$-th column of $A$, $A_i^T A_i = 1$. Note that, because $n > m$, $A$ defines a (redundant) frame. Moreover, we assume that $|A_i^T A_j| < 1$ for any $i \neq j$.

A practical algorithm for solving (1) called linearized Bregman iterations (LBI) is derived by solving the optimization problem (3) approximately (Cai, et al., 2009a, 2009b; Yin, et al., 2008). The square error term in Eq. (3) is replaced by its linear approximation $\langle A^T(Au - f), u - u^k\rangle$ around $u^k$ and a proximity term $\tfrac{1}{2\delta}\|u - u^k\|_2^2$ is added to reflect the limited range of validity of the linear approximation. After some algebra the steps (3) and (4) reduce to the following two-step LBI (Cai, et al., 2009a, 2009b; Yin, et al., 2008):

$$v^{k+1} = v^k - A^T(Au^k - f), \tag{5}$$
$$u^{k+1} = \delta\, \operatorname{shrink}(v^{k+1}, \lambda), \tag{6}$$

where $v^k = p^k + u^k/\delta$ and the component-wise operation

$$\text{shrink}(x, \lambda) = \begin{cases} x - \lambda, & \text{if } x > \lambda \\ 0, & \text{if } -\lambda < x < \lambda \\ x + \lambda, & \text{if } x < -\lambda \end{cases}$$

(Elad, Matalon, Shtok, & Zibulevsky, 2007).

The LBI can be naturally implemented by a network of $n$ parallel nodes, Figure 1, an architecture previously proposed to implement LCA (Rozell, et al., 2008). Such a network combines feedforward projections, $A^T$, and inhibitory lateral connections, $-A^TA$, which implement "explaining away" (Pearl, 1988). At every step, each node updates its component of the internal variable, $v$, by adding the corresponding components of the feedforward signal, $A^Tf$, and the broadcast external variable, $-A^TAu$. Then, each node computes the new value of its component in $u$ by shrinking its component in $v$. Another distributed algorithm called RDA (Xiao, 2010) can also be implemented by such a network.

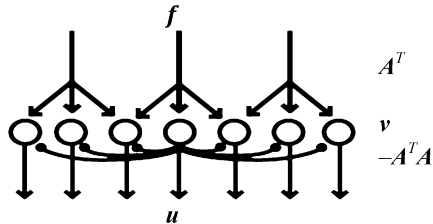

Figure 1: A network architecture for LCA, RDA, LBI, or HDA. Feedforward projections multiply the input $f$ by a matrix $A^T$, while lateral connections update internal node activity $v$ by a product of matrix $-A^TA$ and external activity $u$.

Although LBI, LCA or RDA achieve sparse approximation of the incoming signal, implementing these algorithms in man-made or biological hardware using the network architecture of Fig. 1 would be challenging in practice. The reason is that all these algorithms require real-time communication of analog variables, thus placing high demands on the energy consumption and bandwidth of lateral connections. Considering that the potential number of lateral connections is $O(n^2)$, and both volume and energy are often a limited resource in the brain (Attwell & Laughlin, 2001; Chklovskii, Schikorski, & Stevens, 2002; Laughlin & Sejnowski, 2003) and in sensor networks (Tubaishat & Madria, 2003) we search for a more efficient solution.

## 3    Bregman coordinate descent

In an attempt to find a distributed algorithm for solving (1) under bandwidth limitations, we explore a different strategy, called coordinate descent, where only one component of $\boldsymbol{u}$ is updated at a given iteration (Friedman, et al., 2007; Li & Osher, 2009). Inspired by (Li & Osher, 2009) we derive a novel Bregman coordinate descent algorithm. We start from (3) and rewrite the energy function on the right hand side by substituting matrix notation with explicit summation over vector components:

$$E = \tfrac{1}{2}\left\|\sum_{j=1}^{n} u_j \boldsymbol{A}_j - \boldsymbol{f}\right\|_2^2 + \lambda \sum_{j=1}^{n}\|u_j\|_1 - \lambda \sum_{j=1}^{n}\|u_j^k\|_1 - \sum_{j=1}^{n} p_j^k (u_j - u_j^k). \quad (7)$$

Assuming that in the $(k+1)$-th iteration, the $i$-th component of $\boldsymbol{u}$ is to be updated, and the values of all other components of $\boldsymbol{u}$ remain unchanged, then the updated value $u_i'$ is obtained from

$$u_i' = \operatorname{argmin}_x E = \operatorname{argmin}_x \left\{\tfrac{1}{2}\left\|x\boldsymbol{A}_i + \sum_{j\neq i}^{n} u_j \boldsymbol{A}_j - \boldsymbol{f}\right\|_2^2 + \lambda\|u_i\|_1 - p_i u_i\right\}, \quad (8)$$

where optimization is carried out in respect to the $i$-th component of $\boldsymbol{u}$ denoted by $x$. In iteration (8) we drop terms independent of $u_i$ and do not keep track of the iteration number $k$. The condition for the minimum in (8) is

$$\partial[\lambda\|u_i'\|_1] \ni -\boldsymbol{A}_i^T\left(u_i'\boldsymbol{A}_i + \sum_{j\neq i}^{n} u_j \boldsymbol{A}_j - \boldsymbol{f}\right) + p_i, \quad (9)$$

where $\partial[.]$ designates a subdifferential (Boyd & Vandenberghe, 2004). Noticing $\boldsymbol{A}_i^T \boldsymbol{A}_i = 1$ and $\sum_{j\neq i}^{n} u_j \boldsymbol{A}_j = \boldsymbol{A}\boldsymbol{u} - u_i \boldsymbol{A}_i$, we rewrite (9) as

$$\partial[\lambda\|u_i'\|_1] \ni -u_i' + u_i - \boldsymbol{A}_i^T(\boldsymbol{A}\boldsymbol{u} - \boldsymbol{f}) + p_i, \quad (10)$$

From the optimality condition (10), we get the update formula of $p_i$ (Yin et al., 2008),

$$p_i' + u_i' = p_i + u_i - \boldsymbol{A}_i^T(\boldsymbol{A}\boldsymbol{u} - \boldsymbol{f}), \quad (11)$$

where $\partial[\lambda\|u_i'\|_1] \ni p_i'$. By defining $v_i = p_i + u_i$, we get:

$$v_i' = v_i - \boldsymbol{A}_i^T(\boldsymbol{A}\boldsymbol{u} - \boldsymbol{f}). \quad (12)$$

Then we derive the update formula of $u_i$. Noticing

$$\left\|x\boldsymbol{A}_i + \sum_{j\neq i}^{n} u_j \boldsymbol{A}_j - \boldsymbol{f}\right\|_2^2 = x^2 + 2x\boldsymbol{A}_i^T\left(\sum_{j\neq i}^{n} u_j \boldsymbol{A}_j - \boldsymbol{f}\right) + \left\|\sum_{j\neq i}^{n} u_j \boldsymbol{A}_j - \boldsymbol{f}\right\|_2^2$$

$$= \left\|x + \boldsymbol{A}_i^T\left(\sum_{j\neq i}^{n} u_j \boldsymbol{A}_j - \boldsymbol{f}\right)\right\|_2^2 - \left\|\boldsymbol{A}_i^T\left(\sum_{j\neq i}^{n} u_j \boldsymbol{A}_j - \boldsymbol{f}\right)\right\|_2^2 + \left\|\sum_{j\neq i}^{n} u_j \boldsymbol{A}_j - \boldsymbol{f}\right\|_2^2$$

$$= \left\|x + \boldsymbol{A}_i^T\left(\sum_{j\neq i}^{n} u_j \boldsymbol{A}_j - \boldsymbol{f}\right)\right\|_2^2 + const, \quad (13)$$

we rewrite (8) as

$$u'_i = \operatorname{argmin}_x \left\{ \tfrac{1}{2} \| x + A_i^T (\textstyle\sum_{j \neq i}^n u_j A_j - f) \|_2^2 + \lambda \|u_i\|_1 - p_i u_i \right\}$$
$$= \operatorname{argmin}_x \left\{ \tfrac{1}{2} \| x - p_i + A_i^T (\textstyle\sum_{j \neq i}^n u_j A_j - f) \|_2^2 + \lambda \|u_i\|_1 \right\}$$
$$= \operatorname{shrink}(p_i - A_i^T (\textstyle\sum_{j \neq i}^n u_j A_j - f), \lambda)$$
$$= \operatorname{shrink}(p_i + u_i - A_i^T(Au - f), \lambda). \tag{14}$$

By substituting Eqns. (11) and (12) into (14), we get:

$$u'_i = \operatorname{shrink}(v'_i, \lambda). \tag{15}$$

These iterations appear similar to that in LBI (5, 6), but are performed in a component-wise manner resulting in the following algorithm.

**Algorithm 1: Bregman coordinate descent**
Initialize: $v=0$, $u=0$
**while** "$\|f - Au\|_2^2$ not converge" **do**
    "choose $i \in \{1:n\}$"
$$v_i \leftarrow v_i - A_i^T(Au - f), \tag{16}$$
$$u_i \leftarrow \operatorname{shrink}(v_i, \lambda). \tag{17}$$
**end while**

In addition to specifying component-wise iterations in Algorithm 1, we must also specify the order in which the components of $u$ are updated. Previous proposal include updating components sequentially based on the index $i$ (Friedman, et al., 2007; Genkin, Lewis, & Madigan, 2007), randomly, or based on the gradient of the objective function (Li & Osher, 2009). In general, choosing $i$ in a distributed architecture requires additional communication between nodes and, therefore, places additional demands on energy consumption and communication bandwidth.

## 4    Derivation of the Hybrid Distributed Algorithm (HDA)

Here, we present our central contribution, a distributed algorithm for solving (1), which has lower bandwidth requirements for lateral connections than the existing ones and does not require additional communication for determining the update order. We name our algorithm Hybrid Distributed Algorithm (HDA) because it combines a gradient-descent-like update of $v$, as in Eq. (5), and a coordinate-descent-like update of $u_i$, as in Eq. (17). The key to this combination is the quantization of the external variable, arising from replacing the shrinkage operation with thresholding. As a result:
1. Due to quantization of the external variable, communication between nodes requires only low bandwidth and is kept to a minimum.
2. The choice of a component of $u$ to be updated, in the sense of coordinate descent, is computed autonomously by each node.

To reduce bandwidth requirements, instead of communicating the analog variable $u$, HDA nodes communicate a quantized variable $s \in \{-1,0,1\}^n$ to each other. The variable $u$, which solves (1) is obtained from $s$ by averaging it over time: $u = \lambda \bar{s} = \frac{\lambda}{t}\sum_{k=0}^{t} s^k$.

In HDA, components of $s$ are obtained from the internal variable $v$:

$$s \leftarrow \text{threshold}(v, \lambda), \quad (18)$$

where threshold function is component wise,

$$\text{threshold}(x, \lambda) = \begin{cases} 1, & \text{if } x > \lambda \\ 0, & \text{if } -\lambda \leq x \leq \lambda. \\ -1, & \text{if } x < -\lambda \end{cases}$$

An update for the internal variable $v$ is similar to (5) but with substitution of $u$ by $\lambda s$:

$$v \leftarrow v - A^T(\lambda A s - f). \quad (19)$$

Note that in HDA there is no need to explicitly specify the order in which the components of $u$ are updated because the threshold operation (18) automatically updates the components in the order they reach threshold. Updates (18, 19) lead to the following computer algorithm.

**Algorithm 2: Discrete-time HDA**
Initialize: $v$=0, $u$=0, $s$=0, $t$=0.
**while** "$\|f - Au\|_2^2$ not converge" **do**
$\quad t \leftarrow t + 1$
$\quad v \leftarrow v - A^T(\lambda A s - f),$
$\quad s \leftarrow \text{threshold}(v, \lambda),$
$\quad u \leftarrow ((t-1)u + \lambda s)/t.$
**end while**

Although not necessary, precomputing $A^T A$ and $A^T f$ may speed up algorithm execution.

To gain some intuition for Algorithm 2 consider an example, where $f$ is chosen to coincide with some column of $A$, i.e. $f=A_i$. Then the solution of problem (1) must be $u_i=1$, $u_{j\neq i}=0$. Now, let us see how the algorithm computes this solution.

The algorithm starts with $v=0$, $u=0$, $s=0$. Initially, each component $v_j$ changes at a rate of $A_j^T A_i$ and, while the $i$-th component is below the threshold, $u$ stays at 0. Assuming $\lambda \gg 1$, after $\lambda/(A_i^T A_i) = \lambda$ iterations, $v_i$ reaches the threshold $\lambda$ and $s_i$ switches from 0 to 1. At that time, the other components of $v$ are still below threshold, $|v_{j\neq i}| = |\lambda A_{j\neq i}^T f| = |\lambda A_{j\neq i}^T A_i| < \lambda$ and, therefore the components $s_{j\neq i}$ stay at 0. Note that choosing large $\lambda$ guarantees that no more than one component reaches the threshold at any iteration.

Knowing $s$, we can compute the next iteration for $v$ (19), which is $v = \lambda A^T f - A^T(\lambda A_i s_i - f) = \lambda A^T A_i - A^T \lambda A_i + A^T f = A^T f$. Note that the first and the second terms cancelled because the change in $v$ accumulated over previous $\lambda$ iterations is canceled by receiving broadcast $s_i$. Because $s_i$ switches back to 0, $u_i = \lambda \bar{s}_i = \lambda/\lambda = 1$ as required. From this point on, the above sequence repeats itself. The above cancellation maintains $s_{j\neq i} = 0$ and ensures sparsity of the solution, $u_{j\neq i}=0$.

The HDA updates (18, 19) can be immediately translated into the continuous-time evolution of the physical variables $s(t)$ and $v(t)$ in a hardware implementation.

**Continuous-time evolution:**

$$v(t) = \int_0^t A^T[f - \lambda A s(t')]dt' \tag{20}$$

$$s(t) = \text{spike}(v(t), \lambda), \tag{21}$$

where the spike function is component wise,

$$\text{spike}(v_i(t), \lambda) = \begin{cases} \delta_t, & \text{if } v_i(t) = \lambda \\ 0, & \text{if } -\lambda < v_i(t) < \lambda \\ -\delta_t, & \text{if } v_i(t) = -\lambda \end{cases}$$

and $\delta_t$ stands for a Dirac delta function centered at time $t$.

In this continuous-time evolution, the solution to (1) is given by the scaled temporal average $u(t) = \frac{\lambda}{t}\int_0^t s(t')dt'$.

The HDA can be naturally implemented on a neuronal network, Fig 1. Unlike the LCA (Rozell, et al., 2008) and the LBI (Cai, et al., 2009a, 2009b; Yin, et al., 2008), which require neurons continuously communicating graded potentials, the HDA uses perfect, or non-leaky, integrate-and-fire neurons (Dayan & Abbott, 2001; Koch, 1999). Ideal, or non-leaky, integrate-and-fire neurons integrate inputs over time in their membrane voltage, $v$, (20) and fire a unitary action potential (or spike) when the membrane voltage reaches the threshold, $\lambda$, (21). The inputs come from the stimulus, $A^T f$, and from other neurons, via the off-diagonal elements of $-A^T A$. After the spike is emitted, the membrane voltage is reset to zero due to the unitary diagonal elements of $A^T A$. We emphasize that, in discrete-time simulations, the membrane potential of HDA integrate-and-fire neurons after spiking is reset by subtracting the threshold magnitude rather than by setting it to zero (Brette et al., 2007).

Unlike thresholding in the HDA nodes (21), in biological neurons, thresholding is one-sided (Dayan & Abbott, 2001; Koch, 1999). Such discrepancy is easily resolved by substituting each node with two opposing (on- and off-) nodes. In fact, neurons in some brain areas are known to come in two types (on- and off-) (Masland, 2001).

Therefore, the HDA can be used as a model of computation with integrate-and-fire neurons. In the next section, we prove that $u$, a time-average of $s$, which can be viewed as a firing rate, converges to a solution of $f = Au$.

Finally, for the sake of completeness, we propose the following "hopping" version of the HDA, which does not reduce energy consumption of communication bandwidth, yet is convenient for fast implementation on the CPU architecture for the sake of modeling.

**Algorithm 3: hopping HDA**
Initialize: $v=0$, $u=0$, $s=0$, $t=0$.
**While** "$\|f - Au\|_2^2$ not converge" **do**
    $r = \max|v_i|$,
    $j = \text{argmax}_i |v_i|$,
    **if** $r<\lambda$ **then**
        $t_w = \min[(\lambda \, \text{sign}(A_i^T f) - v_i)/(A_i^T f)]$,
        $j = \text{argmin}_i[(\lambda \, \text{sign}(A_i^T f) - v_i)/(A_i^T f)]$,
        $t \leftarrow t + t_w$,
        $s_j \leftarrow \text{sign}(A_j^T f)$
        $v \leftarrow v + t_w A^T f - \lambda s_j A^T A_j$,
        $u_j \leftarrow ((t-1)u_j + s_j)/t$,

      **else**
            $s_j \leftarrow \text{sign}(v_j)$
            $\boldsymbol{v} \leftarrow \boldsymbol{v} - \lambda \boldsymbol{A}^T \boldsymbol{A}_j s_j,$
            $u_j \leftarrow ((t-1)u_j + \lambda s_j)/t,$
      **end if**
**end while**

As before, precomputing $\boldsymbol{A}^T\boldsymbol{A}$ and $\boldsymbol{A}^T\boldsymbol{f}$ may speed up algorithm execution.

The name "hopping HDA" comes from the fact that, instead of waiting for many iterations to reach the threshold, $\lambda$, the algorithm directly determines the component of $\boldsymbol{v}$ which will be the next to reach the threshold and computes the required integration time in $t_w$. Thus, the idea of hopping is similar to the ideas behind LARS (Efron, et al., 2004) and "kicking" (Osher, et al., 2010). When that component of $v$ reaches the threshold, it broadcasts $-\boldsymbol{A}^T\boldsymbol{A}$ to other neurons instantaneously. We note that in practice, several nodes may exceed the threshold at the same time. In this case, we update super-threshold components based on the magnitude of $v_i$ starting with the largest.

## 5    Asymptotic performance guarantees

In this section, we analyze the asymptotic performance of the HDA by proving three theorems. Theorem 1 demonstrates that the HDA can be viewed as taking steps towards the solutions of a sequence of the Lasso problems whose regularizer coefficient decays in the course of iterations. Theorem 2 demonstrates that the representation error decays as *1/t* in the asymptotic limit. Theorem 3 demonstrates that, in the presence of time-varying noise, the representation error in the asymptotic limit decays also as a power of *t*. All the results are proven for the evolution described by Eqns. (20, 21), but can be easily adapted for the discrete-time case.

Importantly, Theorems 1 and 2 together suggest an intuition for why HDA finds a sparse solution. As the solution of a Lasso problem is known to be sparse (Tibshirani, 1996), it may seem possible that solving a sequence of the Lasso problems, as shown in Theorem 1, would yield a sparse solution. Yet, one may argue that, according to Theorem 1, the regularizer coefficient decays in the course of iterations and, because smaller regularization coefficients should yield less sparse solutions, the final outcome may not be sparse. Note, however, that the driving force for the growth of components of $\boldsymbol{u}$ is given by the representation error, which itself shrinks in the course of iterations according to Theorem 2. Because the error decays with the same asymptotic rate as the regularization coefficient we may still expect that the ultimate solution remains sparse. Indeed, such intuition is born out by numerical simulations as will be demonstrated in Section 6.

**Theorem 1:** Define average external variable at time $t$ as $\bar{s}(t) := \frac{1}{t}\int_0^t s(t')dt'$. Then, provided $\|\bar{s}(t)\|_1 \neq 0$, the energy function $E(t) := \|f - \lambda A\bar{s}(t)\|_2^2 + (\lambda^2/t)\|\bar{s}(t)\|_1$ generated by (20,21) decreases monotonically.

**Proof:** To prove this theorem, we consider separately the change in $E(t)$ during the interval between spikes and the change in $E(t)$ during a spike. We define $w := \int_0^t s(t')dt'$, which does not change during the interval between spikes. Then we replace $\bar{s}(t)$ in $E(t)$ by $w/t$ and obtain after simple algebra:

$$dE(t)/dt = \frac{2\lambda}{t^3}w^T A^T(ft - \lambda Aw) - \frac{2\lambda^2}{t^3}\|w\|_1$$

$$= \frac{2\lambda}{t^3}[w^T v(t) - \lambda\|w\|_1]$$

$$= \frac{2\lambda}{t^3}\sum_{i=1}^n [w_i v(t)_i - \lambda|w_i|]. \tag{22}$$

The second equality follows from Eq. (20). Since $|v(t)_i| < \lambda$, if $\|w\|_1 \neq 0$, $dE(t)/dt < 0$. Therefore, during the interval between spikes, $E(t)$ decreases.

If the $i$-th neuron fires a spike at $t$, $|s(t)_i| = 1$ and $s(t)_{j \neq i} = 0$, then the difference between $E(t)$, just after the spike, and $E(t^-)$, just before the spike is given by (notation $t^-$ means arbitrarily close to $t$ from below),

$E(t) - E(t^-)$

$= \|f - \lambda A\bar{s}(t)\|_2^2 + (\lambda^2/t)\|\bar{s}(t)\|_1 - \|f - \lambda A\bar{s}(t^-)\|_2^2 - (\lambda^2/t)\|\bar{s}(t^-)\|_1$

$= \|f - \lambda A\bar{s}(t^-) - \lambda s(t)_i A_i/t\|_2^2 - \|f - \lambda A\bar{s}(t^-)\|_2^2$

$\quad + \frac{\lambda^2}{t}\left(\sum_{j\neq i}|\bar{s}(t^-)_j| + |\bar{s}(t)_i|\right) - \frac{\lambda^2}{t}\left(\sum_{j\neq i}|\bar{s}(t^-)_j| + |\bar{s}(t^-)_i|\right)$

$= -\frac{2\lambda}{t}s(t)_i A_i^T(f - \lambda A\bar{s}(t^-)) + \frac{\lambda^2}{t^2}\|A_i\|_2^2\|s(t)_i\|_2^2 + \frac{\lambda^2}{t}(|\bar{s}(t)_i| - |\bar{s}(t^-)_i|)$

$= -\frac{2\lambda}{t^2}s(t)_i v(t^-)_i + \frac{\lambda^2}{t^2}\|s(t)_i\|_2^2 + \frac{\lambda^2}{t}(|\bar{s}(t^-)_i + s(t)_i/t| - |\bar{s}(t^-)_i|)$

$= \frac{\lambda^2}{t^2}[\|s(t)_i\|_2^2 + |s_i^t|\text{sign}(\bar{s}(t^-)_i s(t)_i) - 2s(t)_i v(t^-)_i/\lambda]. \tag{23}$

In the above equation, we used the relation $\bar{s}(t) = \bar{s}(t^-) + s(t)/t$, which can be written separately for each component as $\bar{s}(t)_i = \bar{s}(t^-)_i + s(t)_i/t$ and $\bar{s}(t)_{j \neq i} = \bar{s}(t^-)_{j \neq i}$ (because $s(t)_{j \neq i} = 0$ ). Since $s(t)_i v(t^-)_i \to \lambda$, $\|s(t)_i\|_2^2 = |s(t)_i| = 1$, $E(t) - E(t^-) \to 0$ when $\text{sign}(\bar{s}(t^-)_i s(t)_i) = 1$ and $E(t) - E(t^-) < 0$ when $\text{sign}(\bar{s}(t^-)_i s(t)_i) = -1$. Therefore, at spike time, $E(t)$ does not increase. Combining (22) and (23) concludes the proof.

Similarly, for the discrete-time HDA, Algorithm 2, it is easy to show that, for sufficiently large $\lambda$, if $\|\bar{s}(t) := (1/t) \sum_{k=0}^{t-1} s^k\|_1 \neq 0$, the sequence $\{E(t) := \|f - \lambda A\bar{s}(t)\|_2^2 + (\lambda^2/t)\|\bar{s}(t)\|_1\}$ generated by Algorithm 2 decreases monotonically.

**Theorem 2:** There exists an upper bound on the representation error, $\|f - \lambda A\bar{s}(t)\|_2$, which decays as O(1/*t*).

**Proof:** In the continuous-time evolution, $v(t)_i = A_i^T \int_0^t [f - \lambda A s(t')]dt'$. Because of the threshold operation, $|v(t)_i| \leq \lambda$ and, therefore,

$$\left| A_i^T \int_0^t [f - \lambda A s(t')]dt' \right| \leq \lambda. \tag{24}$$

Then, assuming that $A$ has full row rank, $\left\| \int_0^t [f - \lambda A s(t')]dt' \right\|_2$ must be also bounded from above. Then, the representation error can be expressed as:

$$\|f - \lambda A\bar{s}(t)\|_2^2 = \frac{1}{t^2} \left\| \int_0^t [f - \lambda A s(t')]dt' \right\|_2^2 \leq \frac{const}{t^2}. \tag{25}$$

Therefore, $\|f - \lambda A\bar{s}(t)\|_2 \leq \frac{const}{t}$, which concludes the proof.

Similar proof can be given for the discrete-time HDA, although with a different constant.

**Theorem 3:** Assume the signal $f$ is subject to time varying noise, i.e. $f(t) = f^0 + \varepsilon(t)$. If $\left\| \int_0^t \varepsilon(t')dt' \right\|_2^2 = O(t^{\alpha<1})$, then $\lim_{t \to \infty} \|f^0 - \lambda A\bar{s}(t)\|_2 = 0$ and there exist some upper bound of $\|f^0 - \lambda A\bar{s}(t)\|_2$, which decays as $t^{-\min(1,1-\alpha)}$.

**Proof:** Because of the threshold operation, $v$ is bounded from above:

$$\|f^0 - \lambda A\bar{s}(t)\|_2^2$$
$$= \frac{1}{t^2} \left\| \int_0^t [f(t') - \varepsilon(t') - \lambda As(t')]dt' \right\|_2^2$$
$$= \frac{1}{t^2} \left\| \int_0^t [f(t') - \lambda As(t')]dt' \right\|_2^2 - \frac{2}{t^2} \langle \int_0^t \varepsilon(t')dt', \int_0^t [f(t') - \lambda As(t')]dt' \rangle.$$

$$+ \frac{2}{t^2}\left\|\int_0^t \boldsymbol{\varepsilon}(t')dt'\right\|_2^2. \tag{26}$$

Using again the fact that $|\boldsymbol{v}(t)_i| = \left|\boldsymbol{A}_i^T \int_0^t [\boldsymbol{f}(t') - \lambda \boldsymbol{A}\boldsymbol{s}(t')]dt'\right|$ is bounded from above, we obtain

$$\|\boldsymbol{f}^0 - \lambda\boldsymbol{A}\bar{\boldsymbol{s}}(t)\|_2^2 \leq \frac{C^2}{t^2} - \frac{2C}{t^2}\left\|\int_0^t \boldsymbol{\varepsilon}(t')dt'\right\|_2 + \frac{2}{t^2}\left\|\int_0^t \boldsymbol{\varepsilon}(t')dt'\right\|_2^2$$
$$= O\bigl(t^{-\min(2, 2-2\alpha)}\bigr) \tag{27}$$

This concludes the proof. Next, we consider several examples of noise.

In the case of $\boldsymbol{f}$ contaminated by the white noise, $\int_0^t \boldsymbol{\varepsilon}(t')dt' = O(\sqrt{t})$, and the representation error converges as $1/\sqrt{t}$.

In the case of static noise where $\boldsymbol{\varepsilon}(t) = \boldsymbol{\varepsilon}$, we obtain:

$$\|\boldsymbol{f}^0 - \lambda\boldsymbol{A}\bar{\boldsymbol{s}}(t)\|_2^2 \leq \frac{C^2}{t^2} - \frac{2C}{t}\|\boldsymbol{\varepsilon}\|_2 + 2\|\boldsymbol{\varepsilon}\|_2^2, \tag{28}$$

which can be used as a stopping criterion in a de-noising application to prevent over-fitting.

## 6 Numerical results

In this section, we report the results of numerical experiments. In the first experiment, we search for sparse representation (1) of synthesized data using HDA. The elements of the matrix $\boldsymbol{A} \in \mathbb{R}^{64 \times 128}$ are chosen from a normal distribution and column-normalized by dividing each element by the $l_2$ norm of its column. For the noiseless case, we construct vector $\boldsymbol{f}$ as $\boldsymbol{A}\boldsymbol{u}^0$, where $\boldsymbol{u}^0 \in \mathbb{R}^{128}$ is generated by randomly selecting $nz = 10$ locations for non-zero entries sampled from a flat distribution between -0.5 and 0.5. Then, we apply the discrete-time HDA (Algorithm 2) using the network (Fig. 1) with 128 nodes. We set the spiking threshold $\lambda = 10$ and obtain a solution, $\boldsymbol{u} = \lambda\bar{\boldsymbol{s}}$, which is compared with $\boldsymbol{u}^0$, Fig. 2.

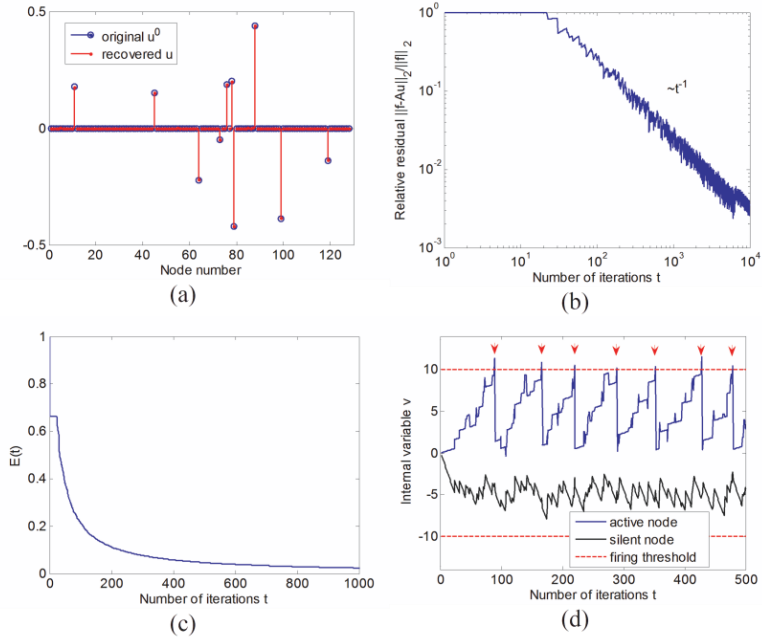

Figure 2: Computing sparse representation, $u$, from noiseless $f = Au^0$ using HDA. (a) The reconstructed $u = \lambda \bar{s}$ (stemmed red dots) at $t = 10000$ coincides with the original $u^0$ (blue circles). (b) The relative residual $\|f - \lambda A\bar{s}\|_2 / \|f\|_2$ decays as $1/t$ (note log-scale axes) in agreement with the upper bound (Theorem 2). The wiggles are due to the discreteness of $s$. (c) Energy, $E^t$, as defined in Theorem 5.1 decays monotonically. (d) Representative evolution of internal variable, $v$, of a broadcasting node (blue) and a silent node (black). Red arrows indicate time points when the component of $s$ corresponding to the broadcasting node is non-zero. The firing thresholds (for $\lambda=10$) are shown by dashed red lines.

As hardware implementations of HDA or neural circuits must operate on the incoming signal $f$ contaminated by noise, which varies during the iterative computation, we analyze the performance of HDA in the presence of noise. To model such a situation we add time varying Gaussian white noise to the original signal $f^0 = Au^0$. On each iteration step, we set each component $f_i^k = f_i^0(1 + 0.5\varepsilon_i^k)$, where the noise $\varepsilon_i^k$ is independently picked from a normal distribution, $N(0,1)$. We found that, despite such a low signal-to-noise ratio, the HDA yields $u$, which is close to the original $u^0$, Fig. 3a. The relative residual decays as $1/\sqrt{t}$, Fig. 4b, as expected from $\sum_{k=1}^{t} \varepsilon^k = O(\sqrt{t})$.

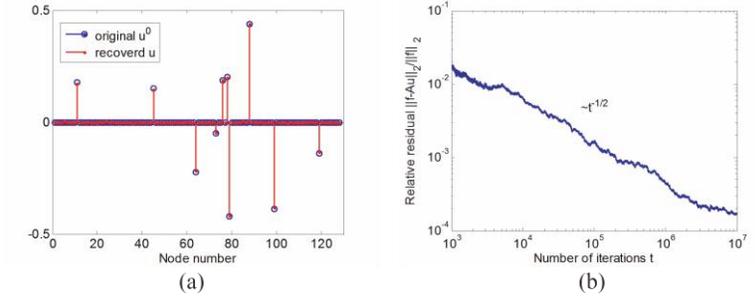

Figure 3: The HDA is robust to noise in the input. Computing sparse representation on the same dataset as Figure 2 but contaminated by strong time-varying noise.

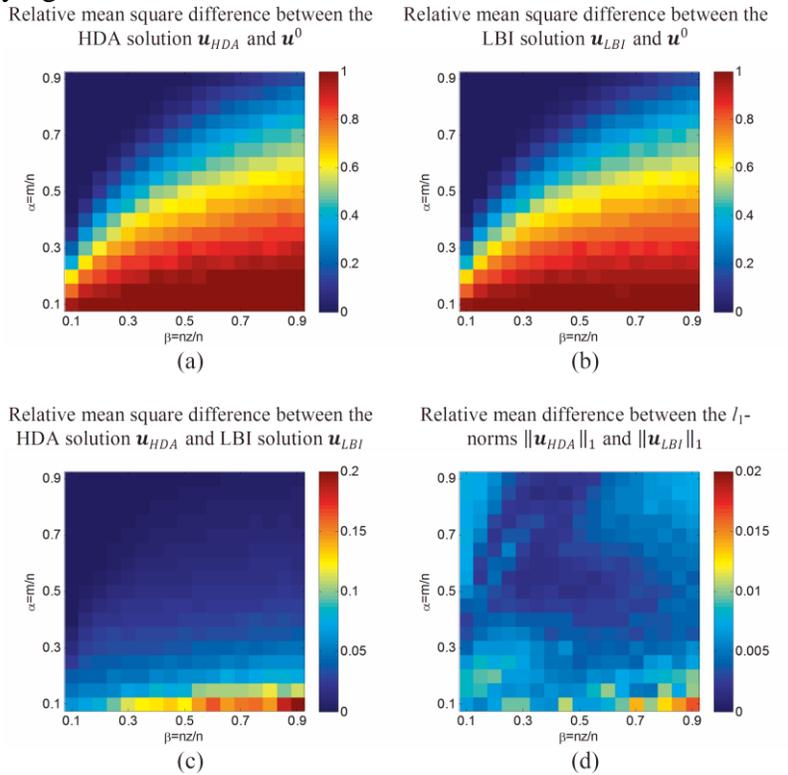

Figure 4: For a wide range of parameters the HDA solution, $u_{HDA}$, is on par with that of the LBI, $u_{LBI}$. The relative mean square difference between $u_{HDA}$ and the predefined sparse signal $u^0$ (a) and the relative mean square difference between $u_{LBI}$ and $u_0$ (b) demonstrate both HDA and LBI both find the unique solution to the basis pursuit problem (1) when it exists (upper left corner). Indeed, the solutions $u_{HDA}$ and $u_{LBI}$ are essentially identical (c) and have the same $l_1$-norms $\|u_{HDA}\|_1$ and $\|u_{LBI}\|_1$ (d).

Next we explore the performance of HDA relative to that of the LBI for a wide range of parameters. We present the results as a function of two variables: system indeterminacy $\alpha = m/n$ and system sparsity $\beta = nz/n$ (Charles, Garrigues, & Rozell, 2011), Fig. 4. We pick $n = 200$ and vary $(\alpha, \beta)$ in the range between 0.1 and 0.9. For each pair $(\alpha, \beta)$, we calculate the corresponding $(m, nz)$ and sample 50 different realizations of the over-complete dictionary $A \in \mathbb{R}^{m \times n}$ and the sparse signal $u^0 \in \mathbb{R}^{200}$ satisfying $\|u^0\|_0 = nz$. We then use HDA and LBI to calculate the corresponding sparse solutions $u_{HDA}$ and $u_{LBI}$. We compare the solution of each algorithm to $u^0$ and plot the relative mean square error $\|u_{HDA/LBI} - u_0\|_2^2 / \|u_0\|_2^2$ in Fig. 4a and b. When the system is sufficiently sparse (small $\beta$) and determinate (large $\alpha$), upper left corners of Fig. 4a and b, the solution to the basis pursuit problem (1) is unique and $u_0$ is perfectly recovered (Chen et al., 1998). Under such condition, the solution of HDA is essentially identical to that of LBI as demonstrated in Fig. 4c, which shows the relative mean square difference between the HDA and the LBI solutions $\|u_{HDA} - u_{LBI}\|_2^2 / \|u_{LBI}\|_2^2$. When $\beta$ gets larger and $\alpha$ gets smaller, the recovery is poor for both algorithms because the predefined $u_0$ is not necessarily the solution with minimum $l_1$-norm and the solution to (1) is not unique (Chen et al., 1998). Therefore the sparse solutions found by HDA and LBI can be very different as revealed by the large difference in the bottom right corner of Fig. 3c, but they still have near identical $l_1$-norms, Fig. 4d. We calculate the relative mean difference between the $l_1$-norms as $\text{abs}(\|u_{LBI}\|_1 - \|u_{HDA}\|_1) / \|u_{LB}\|_1$ and find that the difference averaged over all points in Fig. 4d is only $5 \times 10^{-3}$.

To demonstrate that HDA also serves as model of neural computation, we test it with biologically relevant inputs and dictionary. We use SPAMS (Mairal, Bach, Ponce, & Sapiro, 2010) to train a four times over complete dictionary with 1024 elements from 16×16 image patches randomly sampled form whitened natural images (Olshausen & Field, 1996). These image patches are further processed by subtracting the mean and normalizing contrast by setting variance to unity. The resulting dictionary elements have spatial properties resembling those of V1 receptive fields, Fig. 5a, (Olshausen & Field, 1996).

Then we create a test data set containing 1000 image patches prepared in the same fashion as training image patches. We decompose these image patches using HDA over the learned dictionary and record the mean $l_1$-arc length of the representation coefficients $\|u\|_1$ at various stopping relative residual. As a comparison we also simulate the decompositions using LBI, LCA and RDA. We found that HDA achieves similar representation error − sparsity tradeoff, Fig. 5b and c.

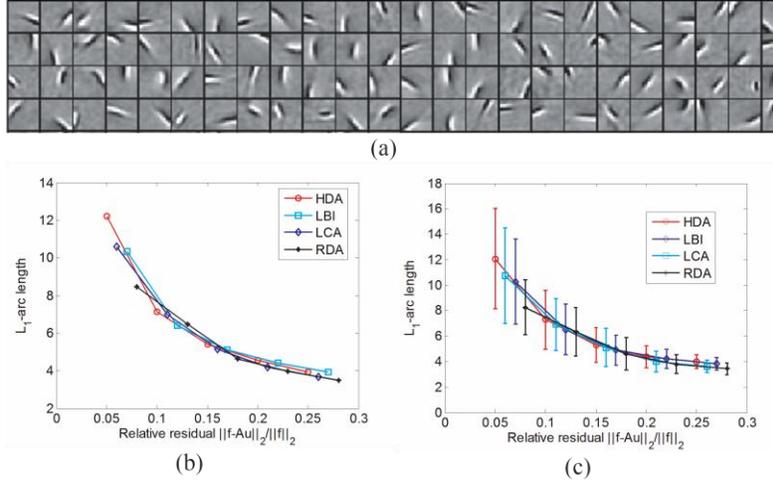

Figure 5: HDA achieves error − sparsity tradeoff comparable with LBI, LCA and RDA. (a) Representative dictionary elements learned from whitened natural image patches. (b) Tradeoff for a typical natural image patch and (c) Mean tradeoff for an ensemble of 1000 contrast normalized image patches.

## 7    Summary

In this paper, we propose an algorithm called HDA, which computes sparse redundant representation using a network of simple nodes communicating using punctuate spikes. Compared to the existing distributed algorithms such as LCA and RDA, the HDA has lower energy consumption and demands on the communication bandwidth. Also, HDA is robust to noise in the input signal. Therefore, HDA is a highly promising algorithm for hardware implementations for energy constrained applications.

We propose three implementations of the HDA: a discrete-time HDA (Algorithm 2), a continuous-time evolution of the physical variable in a hardware implementation, and a hopping HDA (Algorithm 3) for fast computation on a CPU architecture.

Finally, HDA operation combines analog and digital steps (Sarpeshkar, 1998) and is equivalent to a network of non-leaky integrate-and-fire neurons suggesting that it can be used as a model for neural computation.